# Human Following Based on Visual Perception in the Context of Warehouse Logistics

Yanbaihui Liu[1, †], Haibo Wang[2, *, †] and Dongming Jia[3, †]

[1] Department of Electrical & Computer Engineering, University of Michigan, Ann Arbor, U.S.

[2] Faculty of Electronic and Information Engineering, Xi'an Jiao Tong University, Xi'an, China

[3] School of Mechanical and Aerospace Engineering, Ji Lin University, Changchun, China

*Corresponding author: intp0whb@stu.xjtu.edu.cn

†These authors contributed equally to this work

**Abstract.** Warehousing and logistics robots, which have benefited from the development of 5G, the internet, artificial intelligence, and robot technology, are commonly used to assist warehouse personnel in picking up or delivering heavy goods at dispersed locations along dynamic routes. However, traditional programs that can only accept instructions or be preset by the system lack flexibility and existing human auto-following techniques either have difficulty accurately identifying specific targets or rely on a combination of lasers and cameras that are cumbersome and not effective at obstacle avoidance. This paper presents an algorithm that combines DeepSort and a width-based tracking module to track targets and uses artificial potential field local path planning to avoid obstacles. The algorithm is evaluated in a self-designed flat bounded test field and simulated in ROS, and is found to achieve state-of-the-art results in following and successfully reaching the end-point without hitting obstacles.

**Keywords:** Computer vision; Object tracking; Obstacle avoidance.

## 1. Introduction

Currently, in industrial scenarios, most robots refer to the mechanical devices used in the warehousing link and can automatically carry out cargo transfer, handling, and other operations by receiving certain instructions or programs of the system but lack the flexibility and intelligence to meet specific application requirements. For example, in a manufacturing factory scenario, warehouse personnel may need to frequently deliver or retrieve goods from various corners of the production line throughout the day, and pre-programmed handling robots cannot meet this demand for flexibility. Therefore, an automatic following robot that tracks warehouse personnel to deliver goods to a designated location and is also capable of real-time obstacle avoidance is needed.

To solve the problem of the automatic following of a person, how to track the target consistently is the fundamental issue. Many attempts have been made in the past. A large group of them relied on a Laser-based method. Martinez et al. used a 2D laser scanner to do object following and obstacle avoidance [1]. Arras et al. proposed a supervised classifier to detect targets by using 2D lidar data [2] However, due to the limitation of laser, solely relies on it are unable to identify a specific tracking target when multiple similar targets appear. Others either use stereo cameras or combine laser range finders with them. Satake et al. utilize SIFT features and distance and then train an SVM-based person verifier to recognize the specific person [3]. Chen et al. combined modified Online Ada-Boosting with depth information obtained from a stereo camera [4]. To further improve the performance of vision-based methods, Koide et al. proposed a method to determine the target to be focused on using convolutional channel features based on the target predicted by the Unscented Kalman filter given a full body observed [5]. Despite the high level of target tracking and tracing achieved by Koide et al. their method still does not take into account obstacle avoidance, which is a practical problem in real-life factories.

To address the above problems, this paper proposes a solely RGB-D camera-based auto-following system, which enables the robot consistently follow the target person in the factory and avoid hitting any present obstacles. The people-tracking module can deal with multiple people tracking at a close distance, and present switch ID frequently by using DeepSort and a width-based tracking module. The obstacle avoidance module handles the dynamic local path planning so that the robot can avoid stationary machines, non-targeted walking workers, etc., using the Artificial Potential Field (APF). The depth information of the target will be obtained directly from the RGB-D camera instead of being approximated by triangulation techniques since the effect of sunlight on the infrared sensors is minimal in the indoor environment of the factory. Experiments of human following and obstacle avoidance are performed in a flat, bounded test field and simulated in ROS. The results show that our new system can stable track and follow the target person and reach the pre-specific location without hitting obstacles.

## 2. Target detection and path tracking method based on visual perception

### 2.1 Overview

The warehousing environment typically consists of a large number of workers and goods shelves, which requires a robot system to continuously track a specific worker among multiple worker targets and avoid the goods shelves. To accomplish this task, the system needs to be able to perform dynamic object detection and tracking, target following, and obstacle avoidance. The procedure is shown below.

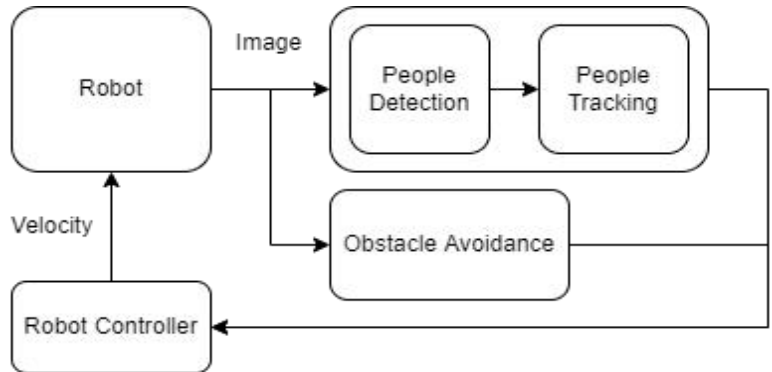

**Fig. 1** Overview of the system

Specifically, we used the depth information of the camera and YOLOv4 to dynamically detect the target [6], then applied the DeepSort algorithm to stably track the moving target [7]. The robot then turns its wheels based on the relative place of the bounding box of the whole image frame and does local path planning for obstacle avoidance by APF [8, 9].

### 2.2 Object detection

In order to improve the real-time performance of target tracking and give consideration to the appropriate tracking accuracy, a tracking algorithm combining YOLOv4 and DeepSort is proposed. This algorithm can not only track and perceive a single target, but also track and perceive multiple targets. You Only Look Once (YOLOv4) is a one-stage object detection model that is well-known for its exceptional computational speed and accuracy and uses CNN once on the entire image to split it into a grid. For each grid location, the CNN network generates a fixed number of different bounding boxes and computes a class probability for each of them. CNN then uses a threshold for the intersections over unions to eliminate bounding boxes that are likely to refer to the same object. This structure of YOLO results in faster computation speed and accurate detection rate. This paper adopted a pre-trained YOLOv4 model on COCO to save time [10]. After successfully detecting the target, DeepSort is performed to do dynamic tracking. The recursive Kalman filtering algorithm is used to predict and track the state of the target candidate box, and then the pedestrian in the continuous multiple frames of the video is tracked and assigned. Three different matching methods, namely, motion matching, appearance matching and cascade matching, are used to achieve more accurate matching. Finally, the continuous tracking of multiple pedestrian targets is achieved, and the ID value of each pedestrian target is obtained in real-time.

### 2.3 Path tracking

To enable the robot to be always oriented toward the target, the target needs to be kept in the center of the camera frame. The center along the horizontal axis (or say orientation in the z-direction) of the bounding box created in the image captured by the camera is used as a reference to move the robot left or right. A threshold range of 20 pixels is used to tolerate errors. The robot turns left whenever it detects any value lesser than the threshold and turns right when greater than the threshold.

For local path planning with obstacle avoidance, we employ an Artificial Potential Field (APF) approach. The APF method involves constructing an artificial virtual potential field, which consists of two components. One part is the gravitational field generated by the target point to the mobile robot, with the direction of the robot pointing to the target point, and the other part is the repulsive field generated by obstacles to the mobile robot, with the direction of the obstacle pointing to the robot. The total potential field in the running space is the combined action of the repulsive force field and gravitational field, so the mobile robot can be controlled by the combined force of gravity and repulsive force.

## 3. Experiment and Analysis

### 3.1 Experimental setup

In the warehouse, there are many goods shelves, and workers walking around, which makes it difficult for robots to follow. However, the layout of the goods shelves in the warehouse is strictly arranged in a rectangular manner, and the roads in the warehouse are relatively open and flat, as shown in Fig.2. We test the multi-person tracking effect of our object detection algorithms with video as input in 3.2. And we use a gazebo to simulate this environment, test two different scenarios and map the motion track of the robot and workers to MATLAB in 3.3 and 3.4.

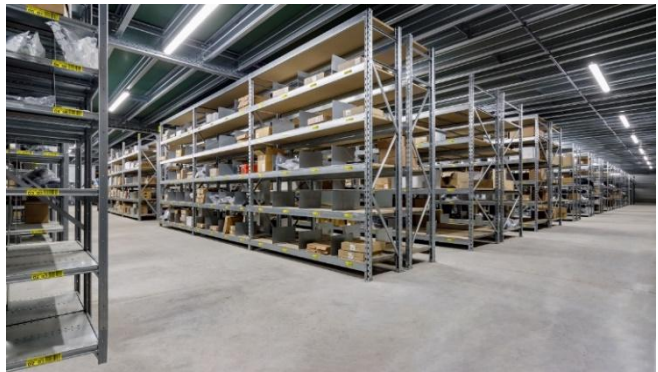

**Fig. 2** Warehouse

### 3.2 Performance evaluation of dynamic target detection and tracking

As for the actual effect of YOLOv4 and DeepSort tracking algorithms, Figure.3 tests the multi-person tracking effect with video as input, mainly to test whether the algorithm can continuously track each pedestrian target in continuous video frames and whether it can keep the ID consistent after workers block each other.

From frames 11 to 60 of Fig.3, it can be seen that although the position of each worker target has changed, the ID of each worker target remains unchanged. This algorithm can continuously track multiple pedestrian targets.

From frames 60 to 115, it can be seen that when workers temporarily occlude each other, the algorithm can accurately track the target, and the IDs of the worker targets do not change before and after occlusion. The algorithms can meet the visual perception requirements of mobile robots for fast and accurate tracking.

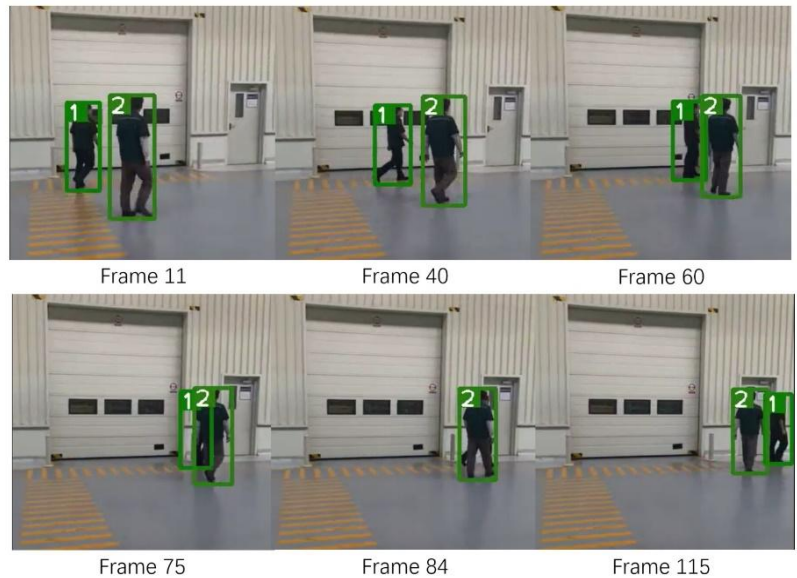

**Fig. 3** Multi-person tracking effect

### 3.3 Analysis of obstacle avoidance performance of right-angle turn

Each area of the warehouse is basically arranged in a rectangular way, so there are many right-angle bends like Fig.4. The following robot has certain deviation from the actual walking path of the target when turning. It is necessary to investigate whether the following robot can move in a small turning radius without collision or target loss. In this experiment, the typical layout gap between goods shelves is used as the running path, and the target is transferred to the channel between two shelves to investigate whether the following robot can successfully follow. Fig.5 is the motion diagram of this experiment. The dotted line in the figure is the walking route of the human body target. The target moves at normal walking speed and turns into an aisle with a small turning radius. It can be seen from Fig.5 that the following robot can successfully follow the target into the equipment aisle even in the case of a small turning radius.

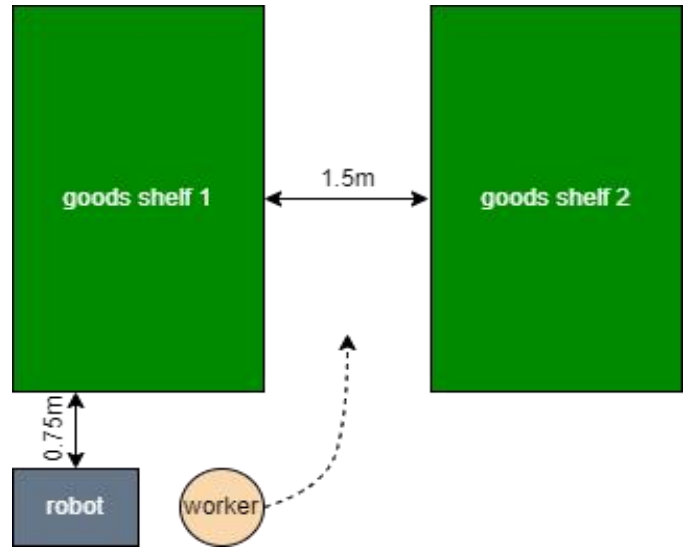

**Fig. 4** Layout

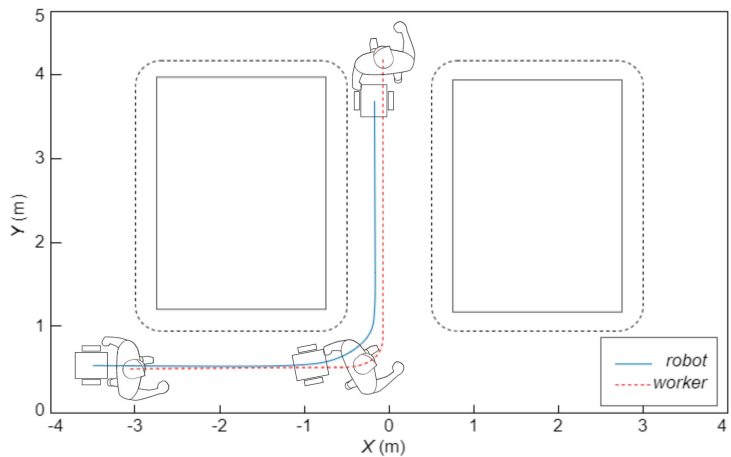

**Fig. 5** Obstacle avoidance performance of right-angle turn

**3.4 Dynamic obstacle avoidance performance analysis**

In addition to avoiding static good shelves, workers walking in the workshop should also be considered when the robot is following. Fig.6 simulates that when a robot follows a worker in the corridor, another worker suddenly walks toward the robot. It can be seen from Fig.6 that the following robot can successfully follow the target when another worker suddenly walks towards the robot.

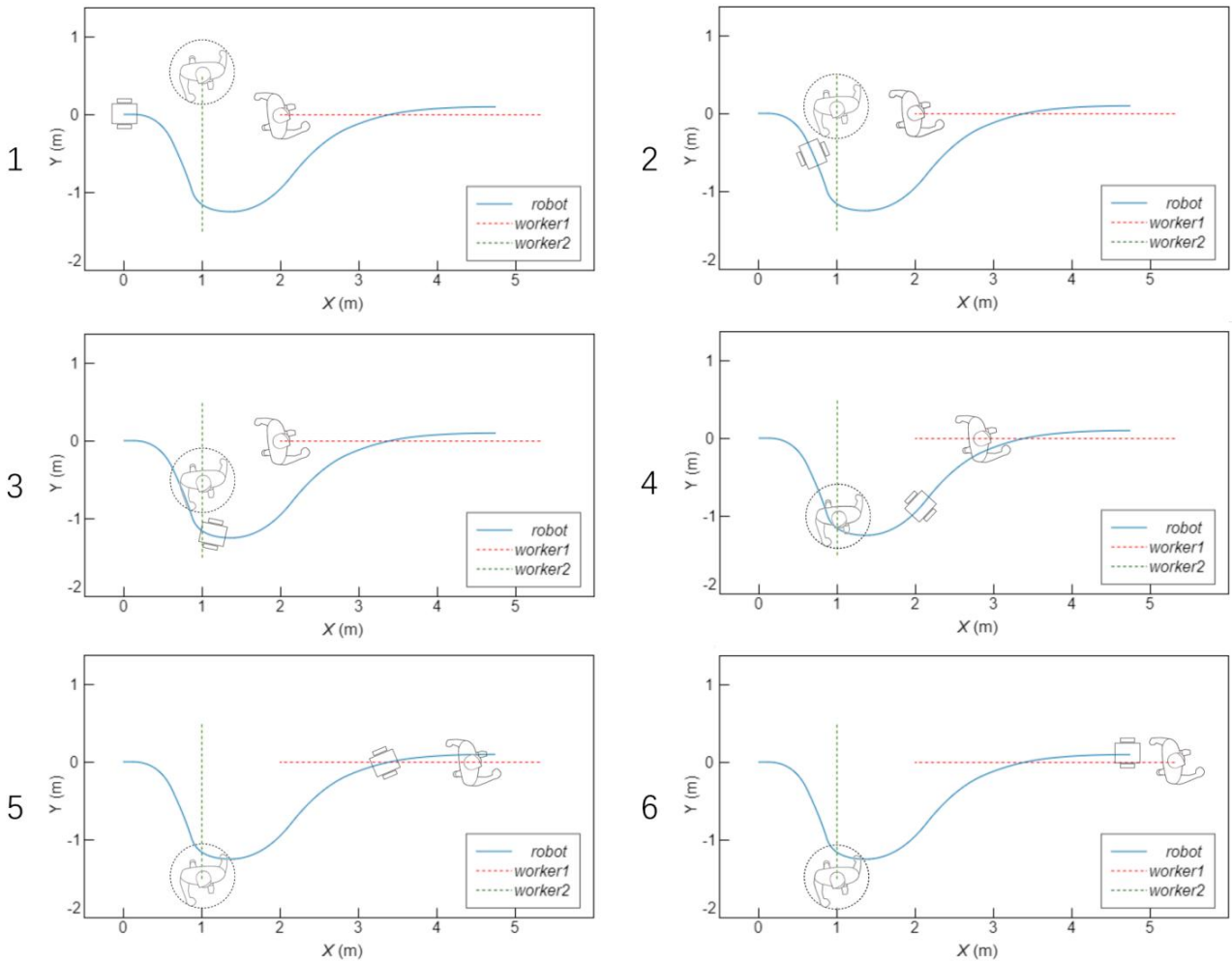

**Fig. 6** Dynamic obstacle avoidance performance

## 4. Conclusion

We successfully accomplish object detection and path tracking based on visual perception in the context of warehousing logistics The system can continue following even if multiple targets temporarily occlude each other, follow the target as it turns at right-angles, and perform well in dynamic obstacle avoidance. If time permits, we will add human pose prediction, which enables the robot to predict the direction of the target in advance and reduce the reaction time.